\documentclass{article}

\usepackage{microtype}
\usepackage{graphicx}
\usepackage{color}
\usepackage{subfigure}
\usepackage{booktabs}
\usepackage{bbm}
\usepackage{multirow}
\usepackage[table,xcdraw]{xcolor}
\usepackage[normalem]{ulem}
\useunder{\uline}{\ul}{}
\usepackage{hyperref}
\usepackage{amsmath}
\usepackage{amssymb}
\usepackage{mathtools}
\usepackage{amsthm}
\usepackage[capitalize,noabbrev]{cleveref}

\usepackage[accepted]{icml2023}

\newcommand{\vx}{x}

\newcommand{\vth}{\theta}
\newcommand{\vt}{\tau}

\theoremstyle{plain}

\theoremstyle{definition}

\theoremstyle{remark}

\usepackage[textsize=tiny]{todonotes}

\icmltitlerunning{Task Addition and Weight Disentanglement in Closed-Vocabulary Models}

\begin{document}

\twocolumn[
\icmltitle{Task Addition and Weight Disentanglement in Closed-Vocabulary Models}

\icmlsetsymbol{equal}{*}

\begin{icmlauthorlist}
\icmlauthor{Adam Hazimeh}{equal,lts4}
\icmlauthor{Alessandro Favero}{equal,lts4,phys}
\icmlauthor{Pascal Frossard}{lts4}
\end{icmlauthorlist}

\icmlaffiliation{lts4}{LTS4, EPFL, Lausanne, Switzerland}
\icmlaffiliation{phys}{PCSL, EPFL, Lausanne, Switzerland}

\icmlcorrespondingauthor{Adam Hazimeh}{adam.hazimeh@epfl.ch}

\icmlkeywords{Machine Learning, ICML}

\vskip 0.3in
]

\printAffiliationsAndNotice{\icmlEqualContribution}

\begin{abstract}
\looseness=-1 Task arithmetic has recently emerged as a promising method for editing pre-trained \textit{open-vocabulary} models, offering a cost-effective alternative to standard multi-task fine-tuning. However, despite the abundance of \textit{closed-vocabulary} models that are not pre-trained with language supervision, applying task arithmetic to these models remains unexplored.
In this paper, we deploy and study task addition in closed-vocabulary image classification models. We consider different pre-training schemes and find that \textit{weight disentanglement} -- the property enabling task arithmetic -- is a general consequence of pre-training, as it appears in different pre-trained closed-vocabulary models. In fact, we find that pre-trained closed-vocabulary vision transformers can also be edited with task arithmetic, achieving high task addition performance and enabling the efficient deployment of multi-task models. Finally, we demonstrate that simple linear probing is a competitive baseline to task addition. Overall, our findings expand the applicability of task arithmetic to a broader class of pre-trained models and open the way for more efficient use of pre-trained models in diverse settings.
\end{abstract}

\section{Introduction}

\looseness=-1 Pre-trained models are widely used as backbones in modern machine learning systems.
However, to enhance their performance on downstream tasks \cite{comprehensive_survey, sanh2021multitask, editing} and increase their robustness \cite{santurkar2021editing, optimism, robust_ft}, these models often require further \textit{editing}. The most common editing method is \textit{fine-tuning}, where pre-trained models are re-trained on specific target tasks. However, aligning models with multiple downstream tasks simultaneously requires joint fine-tuning, which is computationally expensive. 

\looseness=-1 Recently, more cost-effective, scalable, and modular techniques have been introduced, such as editing models directly in weight space via weight interpolation \cite{robust_ft, averaging, rebasin, soups, ties} or task arithmetic \cite{editing, tangent, localizing}. In particular, \textit{task arithmetic} combines multiple, independently fine-tuned model weights through arithmetic operations, thus avoiding the costs of joint fine-tuning. This approach has shown significant potential in preserving both pre-training and fine-tuning performance. However, task arithmetic has thus far only been applied to \textit{open-vocabulary} models, such as CLIP \cite{clip}, which undergo large-scale contrastive pre-training on billions of image-caption pairs and are not limited to a predefined set of classes.

\setcounter{footnote}{1}
In contrast, in computer vision, a major portion of pre-trained models, which we will henceforth refer to as \textit{closed-vocabulary} or \textit{closed}, is not pre-trained with (weak) language supervision but through standard supervised or self-supervised strategies on typically smaller data scales. Importantly, such models lack the flexibility of their open-vocabulary counterparts and require task-specific heads\footnote{For more details on how open-vocabulary models deal with multiple tasks, see Appendix \ref{zs_head}.} depending on the details of the target tasks.  Understanding whether such models can be edited with task arithmetic remains an open question. In particular, \citet{tangent} showed that task arithmetic is enabled by \textit{weight disentanglement} and that such a property emerges with contrastive vision-language pre-training. Yet, as different models can learn varying internal representations, it is unclear if weight disentanglement also emerges with closed-vocabulary pre-training. 

In this paper, we study the scope of task arithmetic to determine whether its success can be leveraged for models pre-trained with diverse pre-training schemes and data scales. Our main contributions are as follows:

\begin{enumerate}
    \itemsep0em 
    \item \looseness=-1 We deploy and study task addition in closed-vocabulary models. In particular, before fine-tuning the encoder, we introduce a task-specific classification head, which we align with the pre-trained encoder via \textit{linear probing}.  
    \item We consider different common pre-training schemes and observe that weight disentanglement is not exclusive to vision-language contrastive pre-training but instead is a general consequence of pre-training.
    \item \looseness=-1 For the same pre-training schemes, we study the performance of task addition with vision transformers of different sizes on 8 image classification tasks, showing that closed-vocabulary models achieve high task addition performance.
    \item Finally, we show that linear probing alone achieves competitive performance to task addition, making it a cheap alternative to task addition for practitioners.
\end{enumerate}

\vspace{-10pt}

\section{Background}

In this section, we present the relevant background on task arithmetic and weight disentanglement. We refer the reader to Appendix \ref{related_work} for additional context.

\paragraph{Task arithmetic}
As introduced by \citet{editing} and further formalized by \citet{tangent}, task arithmetic operates on fine-tuned models by isolating their fine-tuned weights from the pre-trained initialization and performing simple arithmetic operations on the resulting weight differences, known as \textit{task vectors}. Formally, a task $t$ is defined by a dataset and an associated loss function. The corresponding task vector $\tau_t$ is the element-wise difference between the network’s pre-trained weights  $\theta_{\rm pre}$ and its fine-tuned weights $\theta_{\rm ft}^t$, i.e., $\tau_t = \theta_{\rm ft}^t - \theta_{\rm pre}$. Task arithmetic is performed by applying arithmetic operations between different task vectors. For instance, \textit{task addition} adds scaled task vectors to the pre-training weights to produce a multi-task model that is aligned with the target tasks, i.e., $\theta_{\rm new} = \theta_{\rm pre} + \sum_t\lambda_t\tau_t$.

\paragraph{Weight disentanglement}
Such a simple editing technique has shown great performance when applied to open-vocabulary models like CLIP, retaining a significant portion of the accuracy of the single-task fine-tuned models. However, given the non-linear nature of neural networks, why does it work? \citet{tangent} attribute the success of task arithmetic to \textit{weight disentanglement}. This property, seemingly emerging during pre-training, is the ability of a network to decompose its weight space into distinct linear subspaces associated with high performance on different tasks. Weight disentanglement has been observed in \citet{tangent} in different CLIP architectures based on Vision Transformers \cite{vit} and ConvNexts \cite{convnext}. However, it is still unclear if this property can be generalized to \textit{all} pre-training schemes, including those that do not rely on natural language supervision. Furthermore, the influence of different pre-training factors on the emergence of weight disentanglement, such as data characteristics and the training algorithm, is yet to be understood.

To address these gaps, we study task arithmetic, specifically task addition, in the closed-vocabulary setting.

\section{Task Arithmetic with Closed Models}
\label{extending}

\subsection{Handling the Classification Head}
\label{head}
The main challenge in extending task arithmetic to the closed-vocabulary setting is the need for task-specific heads. For open-vocabulary vision models, we can use the model's pre-trained language encoder and leverage the universality of natural language to describe different tasks and classes (see Appendix \ref{zs_head} for more details). However, in the closed setting, this is clearly impossible. 

Following \citet{aligned}, for each task, we propose to first fine-tune a randomly initialized head while freezing the rest of the network -- referred to as \textit{linear probing} -- and then fine-tune the encoder while freezing the head. The first alignment phase is motivated by CLIP fine-tuning, which occurs while keeping the weights of the well-initialized language encoder frozen. In fact, in the closed setting, the first probing phase aligns the head and encoder weights to achieve a better head initialization before fine-tuning the vision encoder. In Appendix \ref{full_ft}, we show that full fine-tuning, i.e., fine-tuning the encoder and a randomly initialized head simultaneously, leads to significantly worse results. Note that, while merging, we apply the task vector to the pre-trained encoder weights and then plug in the task-specific classification head.

\subsection{Experimental Setup}
\label{setup}

\paragraph{Task arithmetic} We focus on task addition over image classification tasks. We consider a uniform set of scaling coefficients for our task vectors, i.e., $\forall t, \; \lambda_t = \lambda$, which is determined via a line search over 21 equispaced values of $\lambda \in \{0, 0.05, 0.1, ..., 0.95, 1 \}$, similarly to \citet{tangent}. The best coefficient is chosen to be the one that maximizes the normalized average accuracy of the resultant model across all tasks, where the normalization is performed with respect to the single-task accuracies of each independently fine-tuned model.

\paragraph{Datasets} We select the same 8 image classification tasks evaluated in \citet{editing}: Stanford Cars \cite{cars}, DTD \cite{dtd}, EuroSAT \cite{eurosat}, GTSRB \cite{gtsrb}, MNIST \cite{lecun-mnisthandwrittendigit-2010}, RESISC45 \cite{resisc45}, SUN397 \cite{sun397}, and SVHN \cite{svhn}. 

\begin{figure*}[ht]
\begin{center}
\centerline{\includegraphics[width=0.95\textwidth]{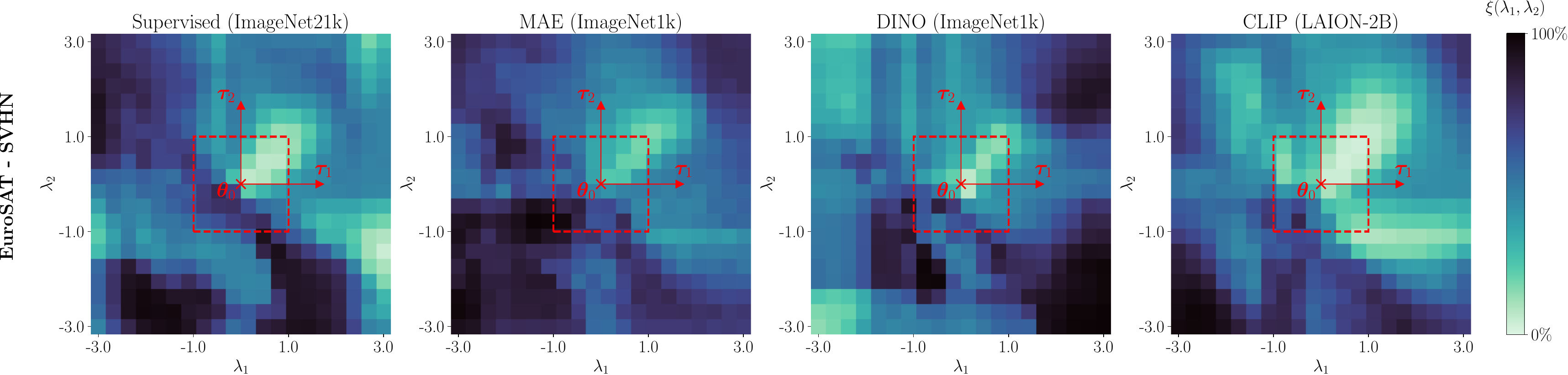}}
\vskip -0.1in
\caption{\textbf{Weight disentanglement error heatmaps for the different pre-training algorithms.} The heatmaps show the pairwise weight disentanglement error $\xi_{\tau_1,\tau_2}(\lambda_1, \lambda_2)$ of ViT-B-16 models pre-trained with different schemes. Light areas denote regions of weight space enjoying stronger weight disentanglement. The red box delimits the search space used to compute the best scaling coefficient $\lambda \in [-1, 1]$.}
\label{wd_main_plot}
\end{center}
\vskip -0.4in
\end{figure*}

\paragraph{Pre-training factors} We study the role of the following pre-training variables:
\begin{enumerate}
    \itemsep0em 
    \item \textit{Pre-training Algorithm:} Supervised, Self-Supervised (Masked Autoencoder \cite{mae} and DINO \cite{dino}), and Contrastive (CLIP). We include the CLIP encoder with a random classification head in order to \textit{i)} directly compare task addition performance to the one obtained by CLIP as reported in \citet{editing} and \citet{tangent}, where it is used in the usual open-vocabulary setting, and \textit{ii)} compare pre-training with and without language supervision.
    \item \textit{Data Size:} ImageNet1k vs. ImageNet21k \cite{imagenet} and LAION-400M \cite{laion400m} vs. LAION-2B \cite{laion5b}, noting that ImageNet and LAION are multiple orders of magnitude apart in terms of the number of samples in each dataset.
    \item \textit{Model Scale:} ViT-B-16 and ViT-L-14/16 \cite{vit}.
\end{enumerate}

We provide a full list of the relevant model checkpoints in Table \ref{checkpoints_tab} of Appendix \ref{checkpoints}.

\section{Results}

In this section, we first study whether closed-vocabulary models also exhibit weight disentanglement, allowing them to perform task arithmetic. Then, we test the effectiveness of task addition with closed-vocabulary models. Finally, we argue that simple baselines, such as linear probing alone, achieve competitive performance.

\subsection{Weight Disentanglement vs. Pre-Training Scheme}
\label{varying}

To investigate the presence of weight disentanglement, we follow the methodology of \citet{tangent} and measure the weight disentanglement error between pairs of tasks $\xi_{\tau_1,\tau_2}(\lambda_1, \lambda_2)$, formally defined as

\vspace{-10pt}

\begin{equation*}
\sum_{t=1}^2\mathbb{E}_{\vx \sim\mu_t} d\left(f(\vx;\vth_{\rm pre}+\lambda_t\vt_t), f(\vx;\vth_{\rm pre}+\lambda_1\vt_1+\lambda_2\vt_2)\right),
\end{equation*}

where $\mu_t$ denotes the input distribution of task $t$, $f(\vx, \theta)$ denotes the output function of the model, and $d$ denotes the prediction error, i.e., $d(y_1,y_2)=\mathbbm1(y_1\neq y_2)$.

Figure \ref{wd_main_plot} displays the disentanglement error for ViT-B-16 models fine-tuned from the different pre-training schemes, for all $\lambda_1, \lambda_2 \in [-3, 3]$. Light regions in the plot indicate areas of the weight space with strong disentanglement. We observe a significant presence of weight disentanglement in regions around the pre-trained initialization, located at the center of the plots, for all models. This finding suggests that weight disentanglement is a general property of pre-training. 

When comparing the disentanglement error of different pre-trained models, we find that those relying on self-supervised pre-training techniques, such as MAE and DINO, achieve less disentanglement compared to models subject to standard supervised or contrastive pre-training. 
Notably, the large-scale pre-training of CLIP achieves the best weight disentanglement among the models considered. Furthermore, in Appendix \ref{wd_extra_plots}, we show that weight disentanglement strengthens as model sizes increase (Figure \ref{wd_plot_model_scale}).

\subsection{Task Addition with Closed-Vocabulary Models}
\label{aligned}

\begin{table*}[ht]
\centering
\caption{\textbf{Task addition performance.} Evaluating the probing, single-task, and task addition accuracy across all 8 tasks (averaged). We vary the data size, training algorithm, and model scale, and indicate the scaling coefficient ($\lambda$) used in task addition for each model.}
\label{aligned_ft_tab}
\vskip 0.15in
\begin{tabular}{@{}cccccc@{}}
\toprule
\begin{tabular}[c]{@{}c@{}}\vspace{-15pt}\textbf{Model}\end{tabular} &
  \multicolumn{2}{c}{\textbf{\begin{tabular}[c]{@{}c@{}}Avg. Single-Task Accuracy (\%)\end{tabular}}} &
  \multicolumn{2}{c}{\textbf{\begin{tabular}[c]{@{}c@{}}Avg. Task Addition Accuracy (\%)\end{tabular}}} &
  \begin{tabular}[c]{@{}c@{}}\vspace{-15pt}$\lambda$\end{tabular} \\
  \cmidrule(lr){2-3}
  \cmidrule(lr){4-5}
 &
  \textit{\textbf{\begin{tabular}[c]{@{}c@{}}Probing\end{tabular}}} &
  \textit{\textbf{Final}} &
  \textit{\textbf{Absolute}} &
  \textit{\textbf{Normalized}} &
   \\ \midrule
\textbf{ViT-B-16}    & \multicolumn{5}{c}{}                                                                            \\
Supervised (IN1k)    & 72.1          & 90.0          & 73.8          & 82.0          & 0.05                        \\
Supervised (IN21k)   & 80.7          & 91.7          & 81.7          & 89.3          & 0.05                        \\
MAE (IN1k)             & 62.8          & 84.5          & 63.1          & 73.0          & 0.05                        \\
DINO (IN1k)            & 82.2          & 90.6          & 82.2          & 90.9          & 0.00 \\
CLIP (LAION-400M)    & \textbf{86.3} & \textbf{92.9} & \textbf{89.8} & \textbf{94.9} & 0.10                        \\
CLIP (LAION-2B)      & 86.0 & 92.6 & 89.6 & 94.5 & 0.15                        \\
\midrule
\textbf{ViT-L-14/16} & \multicolumn{5}{c}{}                                                                            \\
Supervised (IN21k)   & 80.3          & 92.2          & 83.4          & 90.8          & 0.05                        \\
CLIP (LAION-2B)      & \textbf{90.8}          & \textbf{95.8} & \textbf{92.6} & \textbf{96.7} & 0.15                        \\ \bottomrule
\end{tabular}
\vskip -0.1in
\end{table*}

Given the presence of weight disentanglement in closed-vocabulary models, we now turn to the study of task addition as outlined in Section \ref{extending}. Table \ref{aligned_ft_tab} presents the average single-task accuracy alongside the average task addition accuracy (both absolute and normalized\footnote{Normalization is done w.r.t. the single-task (fine-tuning) accuracy, as outlined in Appendix \ref{normalization}.}) for the different pre-trained schemes across all 8 tasks. Notably, task addition performance is high for all models, with the only exception being MAE, which retains only 73\% of the fine-tuning performance of the single-task models.\footnote{Notice that this result aligns with the fact that MAE displays less weight disentanglement, cf. Figure \ref{wd_main_plot}.} The pre-trained CLIP vision encoder significantly outperforms all other models, followed by the ViT model pre-trained on Imagenet21k with standard supervision. Consistently with the weight disentanglement results, models pre-trained on larger pre-training corpora and with larger parameter sizes achieve better task addition performance.

\begin{figure}[t]
\vskip 0.1in
\begin{center}
\centerline{\includegraphics[width=0.9\columnwidth]{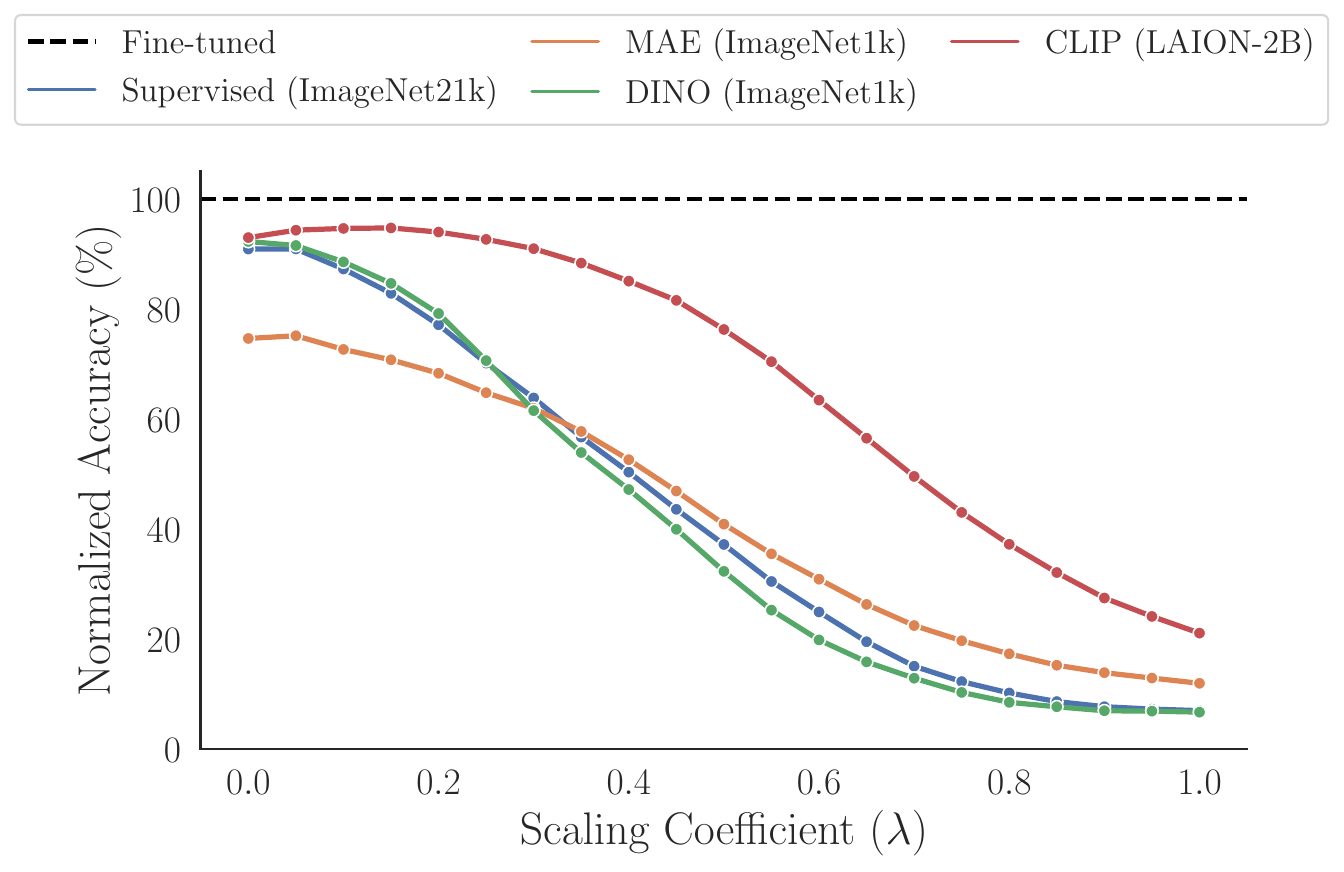}}
\vskip -0.1in
\caption{\textbf{Average normalized task addition accuracy for different pre-training algorithms as a function of the scaling coefficient $\lambda$}. The value $\lambda=0$ corresponds to simple linear probing with no task vector added to the visual encoder.}
\label{aligned_lambda}
\end{center}
\vskip -0.4in
\end{figure}

A closer examination of the scaling coefficients of the task vectors $\lambda$ reveals that their values are small, indicating small changes to the pre-trained visual encoder's weights. This suggests that the high addition performance can be largely attributed to the performance obtained by only probing the head while using the pre-trained encoder weights, implying a minor effect of task addition. Indeed, we notice that task addition accuracies are close to the average single-task accuracies obtained just after linear probing the classification head. Figure \ref{aligned_lambda} shows the task addition normalized accuracies of all pre-trained models while varying the scaling coefficient $\lambda$. This plot indicates that task addition provides only a small (2-3\%) performance gain over linear probing (which corresponds to setting $\lambda=0$), or offers no advantage in the case of DINO pre-training. As $\lambda$ increases beyond the optimal values, normalized accuracy decreases monotonically.

\paragraph{Probing vs. task addition} Linear probing is a significantly cheaper strategy compared to closed-vocabulary task addition, as it only requires training a linear layer for each task while keeping the visual encoder frozen at its pre-trained state. Therefore, at the cost of 2-3 accuracy points, linear probing can be a competitive alternative to task addition in the closed-vocabulary setting for non-critical applications. Moreover, when comparing the average probing performance with that of task addition for open-vocabulary models \cite{editing, tangent}, we find that linear probing achieves considerably higher multi-task performance (see Table \ref{openclip_tab} in Appendix \ref{openclip}). Crucially, as probing only updates the weights of the task-specific heads, the new knowledge acquired with fine-tuning is leveraged solely for those specific target tasks in isolation and with the specific class labels used during single-task fine-tuning. Thus, for users open to forgoing the modularity and flexibility of open-vocabulary models and interested only in high multi-task performance on a stationary set of target tasks, our results show that probing alone can be a valid alternative to task addition.

\section{Conclusion}
In this work, we extended the application of task addition to a variety of closed-vocabulary image classification settings. We showed that weight disentanglement is not exclusive to vision-language contrastive pre-training but is a more general consequence of pre-training. Thus, task arithmetic can be used for editing a much larger class of models. Moreover, we demonstrated that simple linear probing can achieve competitive multi-task accuracy to both closed-vocabulary and open-vocabulary task addition at the expense of sacrificing the open-vocabulary nature of models like CLIP, which was not observed in previous studies.

In the future, understanding if task arithmetic can be applied to models sharing the same backbone (e.g., feature extractor) but differ in the final layers (e.g., image classification vs. segmentation) is an interesting extension of this work. Moreover, given the presence of weight disentanglement in models not undergoing large-scale pertaining such as CLIP, this study opens the avenue for studying the emergence of weight disentanglement during pre-training by performing controlled experiments in smaller-scale scenarios.

\section*{Acknowledgements}
We thank Guillermo Ortiz-Jimenez for the many insightful discussions and his guidance throughout this project. We also thank the anonymous reviewers for their helpful feedback and comments.

\bibliography{references}
\bibliographystyle{icml2023}

\newpage
\appendix
\onecolumn

\section{Further Related Work}
\label{related_work}

Significant progress has been made in studying weight interpolation techniques to enhance the capabilities of pre-trained models. Recent studies demonstrate that interpolating between fine-tuned weights and their pre-trained initializations can improve single-task performance \cite{fisher, ratatouille, robust_ft, averaging, lmc}. Similarly, averaging weights from multiple independently fine-tuned models has been shown to produce high-performing multi-task models \cite{editing, tangent, localizing, ties, patching, soups}. Such techniques can provide better parameter initializations for later training \cite{cold_fusion, fusing}, as well as reduce catastrophic forgetting \cite{forgetting}. Notably, task arithmetic, a multi-task weight interpolation technique, is modular, allowing it to directly accept checkpoints downloaded from online repositories in a plug-and-play fashion. However, notice that successful interpolation requires alignment between the models being merged, ensuring they share a common optimization path early in training \cite{kernel, rebasin}.

\section{Zero-shot Head Initialization}
\label{zs_head}

To initialize a zero-shot head for open-vocabulary models like CLIP, the following approach is typically used:

\begin{enumerate}
    \item For each task, collect its associated set of textual classes (e.g., $C = \{ \textrm{Airplane, \ Car, \ Bird, \ Cat, \ Dog, \ ...}\}$ of size $N$ and specify a set of templates into which the classes can be plugged (e.g., $T = \{ \textrm{A photo of [class], A blurry photo of [class], A photo of a big [class], ...} \}$
    \item Use the model’s text encoder to obtain embeddings of dimension $d_{\textrm{emb}}$ for each class and each template.
    \item Compute the average of these embeddings across each class to obtain a general text embedding of all classes $W_{\rm text} \in \mathbb{R}^{d_{\textrm{emb}} \times N}$
    \item Build an empty classification head and initialize its weights using $W_{\textrm{text}}$, connecting it to the model’s image encoder.
    \item Discard the text encoder and fine-tune the image encoder only (freezing the zero-shot head).
\end{enumerate}

\section{Experimental Hyperparameters}
\label{hparams}
We follow the same hyperparameter setting as \citet{tangent} and \citet{editing}. Namely, for each model configuration, we fine-tune all datasets starting from the same model checkpoint. We fine-tune (and probe when applicable) for 2,000 iterations with a batch size of 128 using the AdamW optimizer \cite{adamw}, and a learning rate determined by line search on the Stanford Cars \cite{cars} dataset (Probing: \{1e-3, 3e-3, 1e-2, 3e-2, 1e-1, 3e-1\} and full or encoder fine-tuning: \{1e-6, 1e-5, 1e-4, 1e-3, 1e-2, 1e-1\}) under a cosine annealing learning rate schedule with 200 warmup steps.

\section{Model Checkpoints}
\label{checkpoints}
In Table \ref{checkpoints_tab}, we list the model checkpoints used in this study.

\begin{table}[h]
\centering
\caption{\textbf{HuggingFace Model Checkpoints.} The URL of each checkpoint is also provided as a hyperlink.}
\label{checkpoints_tab}
\vskip 0.2in
\begin{tabular}{@{}ccc@{}}
\toprule
\textbf{Pre-training Scheme}                  & \textbf{Architecture} & \textbf{HuggingFace Repository}                                                                       \\ \midrule
                                     & ViT-B-16     & \href{https://huggingface.co/timm/vit_base_patch16_224.augreg_in21k}{timm/vit\_base\_patch16\_224.augreg\_in21k}                                                  \\
\multirow{-2}{*}{Supervised (IN21k)} & ViT-L-16     & \href{https://huggingface.co/google/vit-large-patch16-224-in21k}{google/vit-large-patch16-224-in21k}                                                           \\
Supervised (IN1k)                    & ViT-B-16     & \href{https://huggingface.co/timm/vit_base_patch16_224.augreg_in1k}{timm/vit\_base\_patch16\_224.augreg\_in1k}                       \\
MAE (IN1k)                           & ViT-B-16     & \href{https://huggingface.co/facebook/vit-mae-base}{facebook/vit-mae-base}                                                                        \\
DINO (IN1k)                          & ViT-B-16     & \href{https://huggingface.co/facebook/dino-vitb16}{facebook/dino-vitb16}                                                                         \\
CLIP (LAION-400M)                    & ViT-B-16     & \href{https://huggingface.co/laion/scaling-laws-openclip/tree/main}{laion/Model-B-16\_Data-400M\_Samples-34B\_lr-1e-3\_bs-88k} \\
CLIP (LAION-2B)                      & ViT-B-16     & \href{https://huggingface.co/laion/scaling-laws-openclip/tree/main}{laion/Model-B-16\_Data-2B\_Samples-34B\_lr-1e-3\_bs-88k}   \\ \bottomrule
\end{tabular}
\end{table}

\section{Normalized Task Addition Accuracy}
\label{normalization}
The normalization of the task addition accuracy is done with respect to the average single-task accuracy obtained by independently fine-tuning on each task. In particular,

\begin{equation}
    \text{Normalized accuracy}=\cfrac{1}{T}\sum_{t=1}^T\frac{\underset{\vx\sim\mu_t}{\operatorname{acc}}\left[f(\vx;\vth_{\rm pre}+\sum_{t'} \vt_{t'})\right]}{\underset{\vx\sim\mu_t}{\operatorname{acc}}\left[f(\vx;\vth_{\rm pre}+\vt_t)\right]}.
\end{equation}

\section{Further Weight Disentanglement Results}
\label{wd_extra_plots}

\subsection{Effect of Pre-training Data Size}
\label{data_size}
\begin{figure}[h]
\vskip 0.2in
\begin{center}
\centerline{\includegraphics[width=300pt]{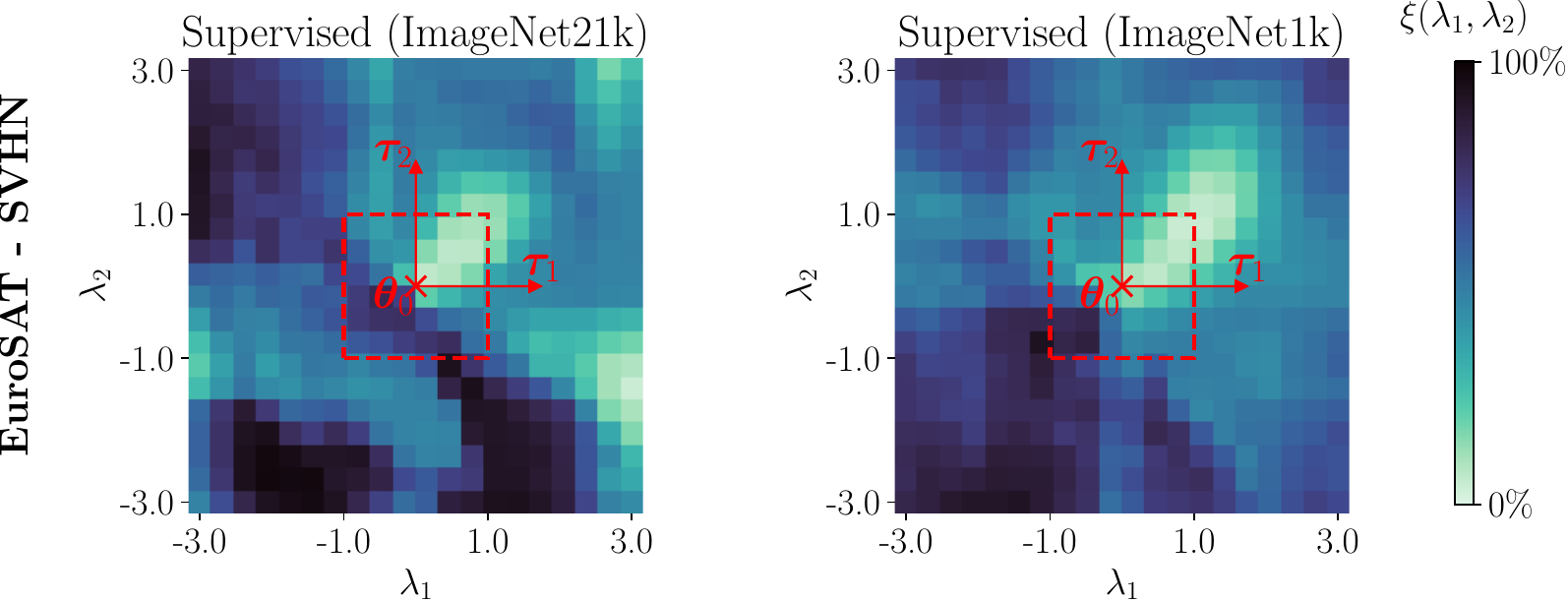}}
\vskip -0.1in
\caption{\textbf{Weight disentanglement error heatmaps for Supervised pre-training on ImageNet1k vs. ImageNet21k.} Both models are based on a ViT-B-16 architecture. The red box delimits the search space used to compute the best scaling coefficient $\lambda$.}
\label{wd_plot_data_size}
\end{center}
\end{figure}

We compare the weight disentanglement error of a Supervised ViT-B-16 pre-trained on ImageNet1k vs. ImageNet21k ($\sim$1 order of magnitude difference in data size) in Figure \ref{wd_plot_data_size}.
While the observed gain in WD strength might seem minimal, the results in Tables \ref{aligned_ft_tab} and \ref{full_ft_tab} indicate that the ImageNet21k model can achieve higher task addition accuracy compared to the ImageNet1k ViT.

\subsection{Effect of Model Scale}
Similar to Appendix \ref{data_size}, we compare the error of two supervised models with the same pre-training dataset (ImageNet21k), but with a varying number of parameters (ViT-B-16 vs. ViT-L-16). We can clearly observe that the larger ViT-L-16 is significantly more weight disentangled than its smaller counterpart. This finding is supported by the results in Tables \ref{aligned_ft_tab} and \ref{full_ft_tab}, which show that the supervised ViT-L-16 admits a task addition advantage over ViT-B-16.
\begin{figure}[h]
\vskip 0.2in
\begin{center}
\centerline{\includegraphics[width=300pt]{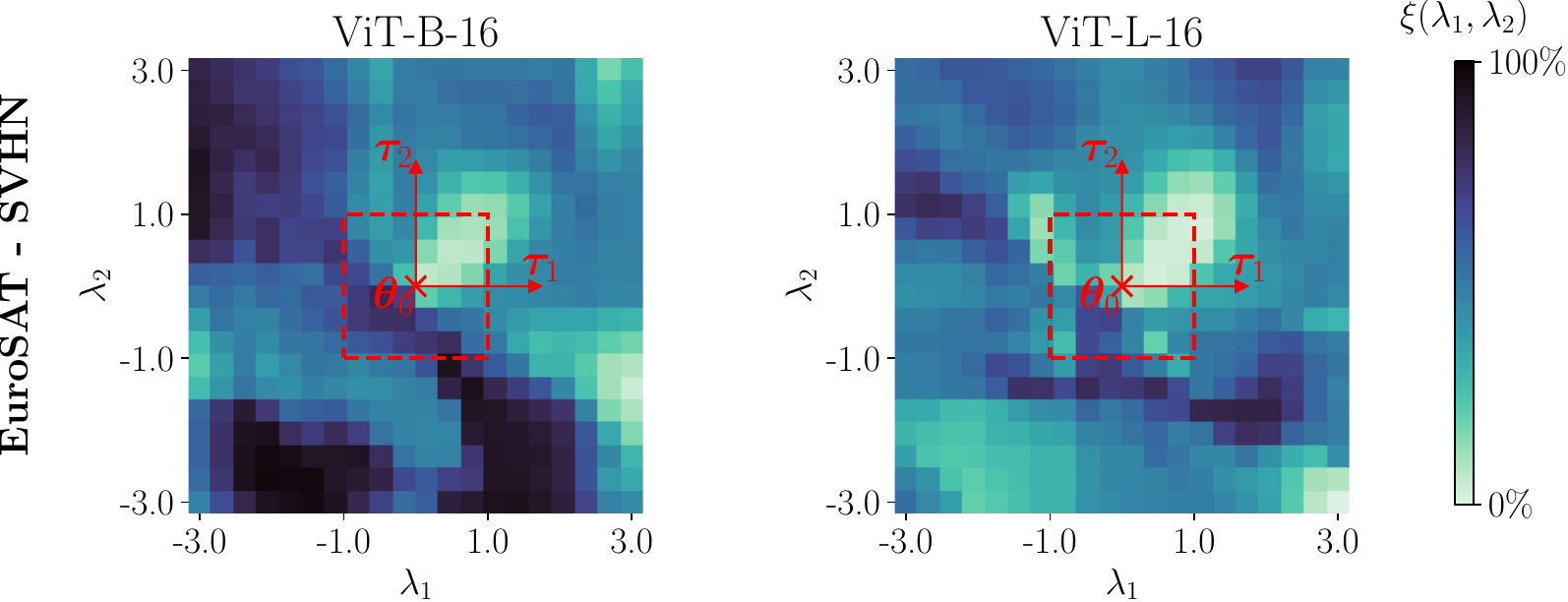}}
\vskip -0.1in
\caption{\textbf{Weight disentanglement error heatmaps highlighting the effect of model scale.} The two models are ViT-B and ViT-L, both pre-trained on ImageNet21k in a supervised manner. The red box delimits the search space used to compute the best scaling coefficient $\lambda$.}
\label{wd_plot_model_scale}
\end{center}
\vskip -0.1in
\end{figure}

\section{Full Fine-tuning}
\label{full_ft}
Table \ref{full_ft_tab} reports the task addition performance under the full-fine tuning regime, wherein both the image encoder and the classification head are simultaneously fine-tuned. While task addition achieves non-trivial accuracy with full fine-tuning, it significantly loses downstream performance when comparing each model's absolute task addition accuracy to its average single-task accuracy, as opposed to the results obtained with aligned fine-tuning (Table \ref{aligned_ft_tab}). Notably, if we consider full fine-tuning for CLIP (LAION-2B) in the \textit{open} setting (Table \ref{openclip_tab}; starting from CLIP's zero-shot head initialization), we can see that it achieves around 72\% absolute addition accuracy for ViT-B-16, compared to 47\% in the closed setting (with similar single-task performance). This suggests that the full fine-tuning approach might be inadequate to fully leverage the potential of closed-task arithmetic.

\begin{table*}[ht]
\centering
\caption{\textbf{Full Fine-tuning: Evaluating the single-task and task addition accuracy across all tasks (averaged).} We vary the data size, training algorithm, and model scale. We also indicate the scaling coefficient ($\lambda$) used in task addition as done in Table \ref{aligned_ft_tab}.}
\label{full_ft_tab}
\vskip 0.15in
\begin{tabular}{@{}ccccc@{}}
\toprule
\begin{tabular}[c]{@{}c@{}}\vspace{-15pt}\textbf{Model}\end{tabular} &
  \textbf{\begin{tabular}[c]{@{}c@{}}\vspace{-15pt}Avg. Single-Task Accuracy (\%)\end{tabular}} &
  \multicolumn{2}{c}{\textbf{\begin{tabular}[c]{@{}c@{}}Avg. Task Addition Accuracy (\%)\end{tabular}}} &
  \begin{tabular}[c]{@{}c@{}}\vspace{-15pt}$\lambda$\end{tabular} \\
  \cmidrule(lr){3-4}
                     &       & \textit{\textbf{Absolute}} & \textit{\textbf{Normalized}} &      \\ \midrule
\textbf{ViT-B-16}    & \multicolumn{4}{c}{}                                                     \\
Supervised (IN1k)    & 87.9 & 44.5                      & 46.5                        & 0.35 \\
Supervised (IN21k)   & 90.2 & \textbf{55.5}                      & \textbf{58.5}                        & 0.25 \\
MAE (IN1k)             & 82.1 & 25.3                      & 26.9                        & 0.25 \\
DINO (IN1k)            & 87.9 & 27.9                      & 28.8                        & 0.25 \\
CLIP (LAION-400M)      & 91.6 & 45.8                      & 47.1                        & 0.30 \\
CLIP (LAION-2B)      & \textbf{92.0} & 47.0                      & 48.2                        & 0.30 \\
\midrule
\textbf{ViT-L-14/16} &       &                            &                              &      \\
Supervised (IN21k)   & 90.9 & 59.9                      & 62.2                        & 0.25 \\
CLIP (LAION-2B)      & \textbf{94.1} & \textbf{70.9}                      & 72.9                        & 0.40 \\ \bottomrule
\end{tabular}
\end{table*}

\begin{figure*}[ht]
\vskip 0.2in
\begin{center}
\centerline{\includegraphics[width=\textwidth]{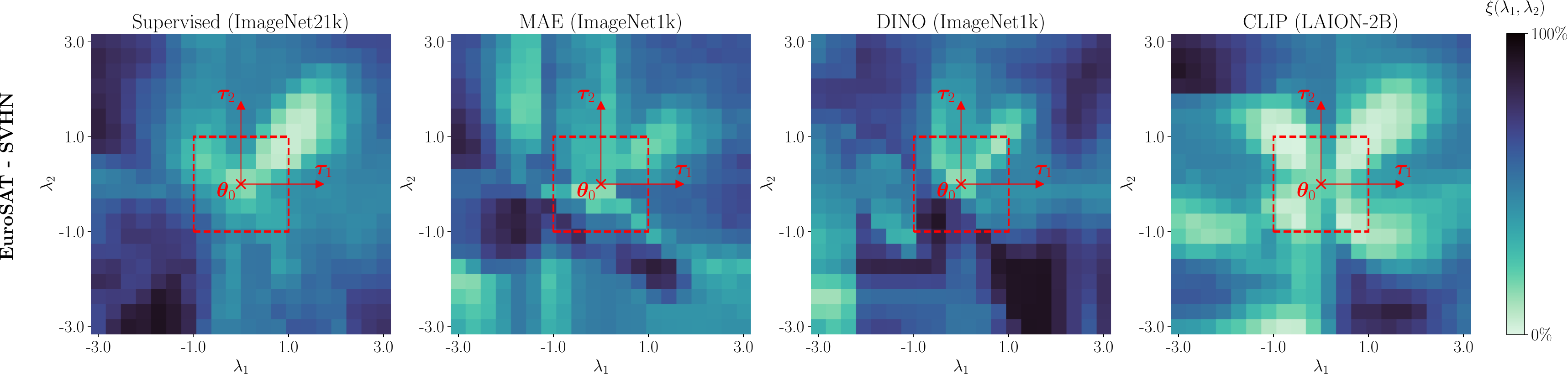}}
\caption{\textbf{Full Fine-tuning: Weight disentanglement error heatmaps for the different pre-training algorithms.} All models are based on a ViT-B-16 architecture following the full fine-tuning regime. The red box delimits the search space used to compute the best scaling coefficient $\lambda$.}
\label{wd_plot_full_ft}
\end{center}
\end{figure*}

\section{Task Addition with Open-vocabulary CLIP}
\label{openclip}
We perform task addition on open-vocabulary CLIP under two different fine-tuning regimes: Encoder fine-tuning (as done in \citet{editing} and \citet{tangent}, wherein the classification head is frozen) and full fine-tuning. We report the results in Table \ref{openclip_tab}.

\begin{table}[h]
\centering
\caption{\textbf{Open CLIP Fine-tuning.} Task addition on open-vocabulary CLIP (ViT-B-16 on LAION-2B).}
\label{openclip_tab}
\vskip 0.15in
\begin{tabular}{@{}cccccc@{}}
\toprule
\textbf{\begin{tabular}[c]{@{}c@{}}\vspace{-15pt}Fine-tuning Regime\end{tabular}} &
  \multicolumn{2}{c}{\textbf{\begin{tabular}[c]{@{}c@{}}Average Single-Task\\ Accuracy (\%)\end{tabular}}} &
  \multicolumn{2}{c}{\textbf{\begin{tabular}[c]{@{}c@{}}Average Task Addition\\ Accuracy (\%)\end{tabular}}} &
  \begin{tabular}[c]{@{}c@{}}\vspace{-15pt}$\lambda$\end{tabular}\\
  \cmidrule(lr){2-3}
  \cmidrule(lr){4-5}
        & \textit{\textbf{Zeroshot}} & \textit{\textbf{Final}} & \textit{\textbf{Absolute}} & \textit{\textbf{Normalized}} &      \\ \midrule
Encoder & 54.42                      & 91.36                   & 72.36                      & 77.21                        & 0.25 \\
Full    & 54.42                      & 91.63                   & 71.16                      & 75.89                        & 0.25 \\ \bottomrule
\end{tabular}
\end{table}

\section{Varying the Number of Tasks: Extra Experiments}

\begin{figure}[h]
\vskip 0.2in
\begin{center}
\centerline{\includegraphics[width=300pt]{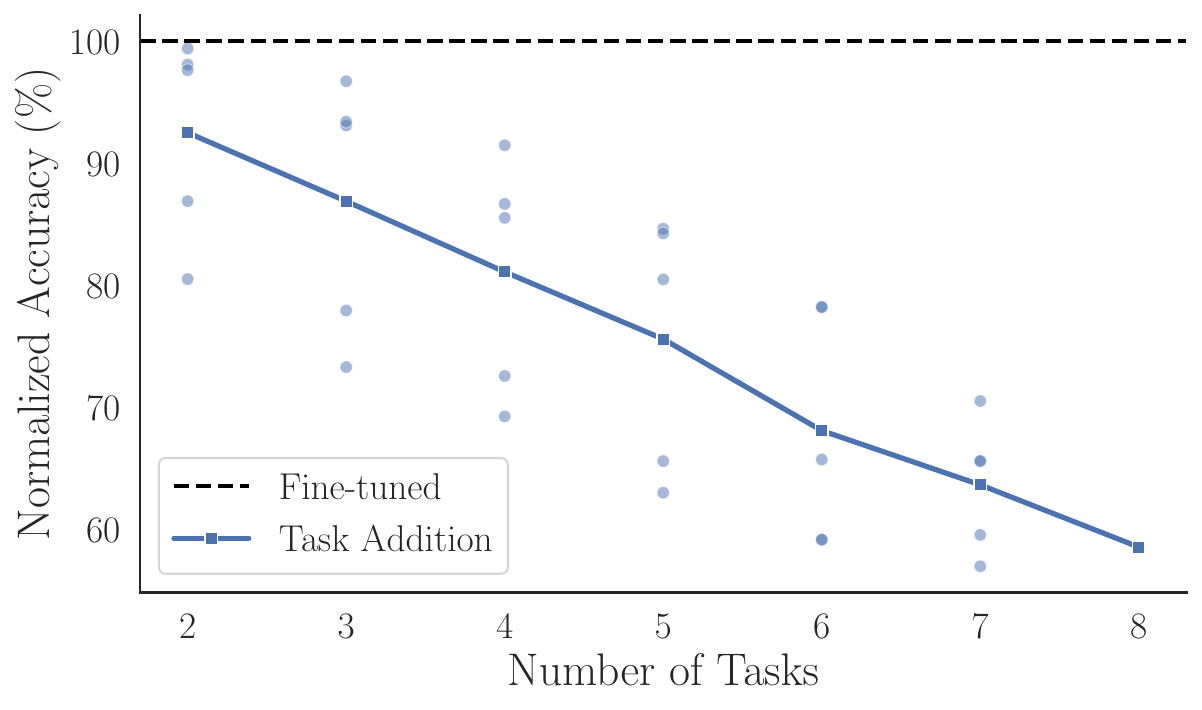}}
\vskip -0.1in
\caption{\textbf{Task addition performance for Supervised ViT-B-16 (ImageNet21k) under a varying number of tasks.} The solid line represents the mean accuracy, while the dots represent different combinations of tasks.}
\label{vary_combinations}
\end{center}
\end{figure}

Our results in Figure \ref{wd_plot_full_ft} reveal a strong presence of weight disentanglement in a region around the pre-trained initialization of models that follow the full fine-tuning regime. This finding suggests that the suboptimal performance of task addition under this regime might be a direct result of evaluating too many tasks. To verify this claim,  we evaluate task addition while varying the number of merged tasks. In Figure \ref{vary_combinations}, we plot task addition performance for a Supervised ViT-B-16 pre-trained ImageNet21k and multiple different combinations of tasks. We observe that performance is high when the number of tasks is small and follows linear decay with the number of tasks. 

\begin{figure}[h]
\vskip 0.2in
\begin{center}
\centerline{\includegraphics[width=300pt]{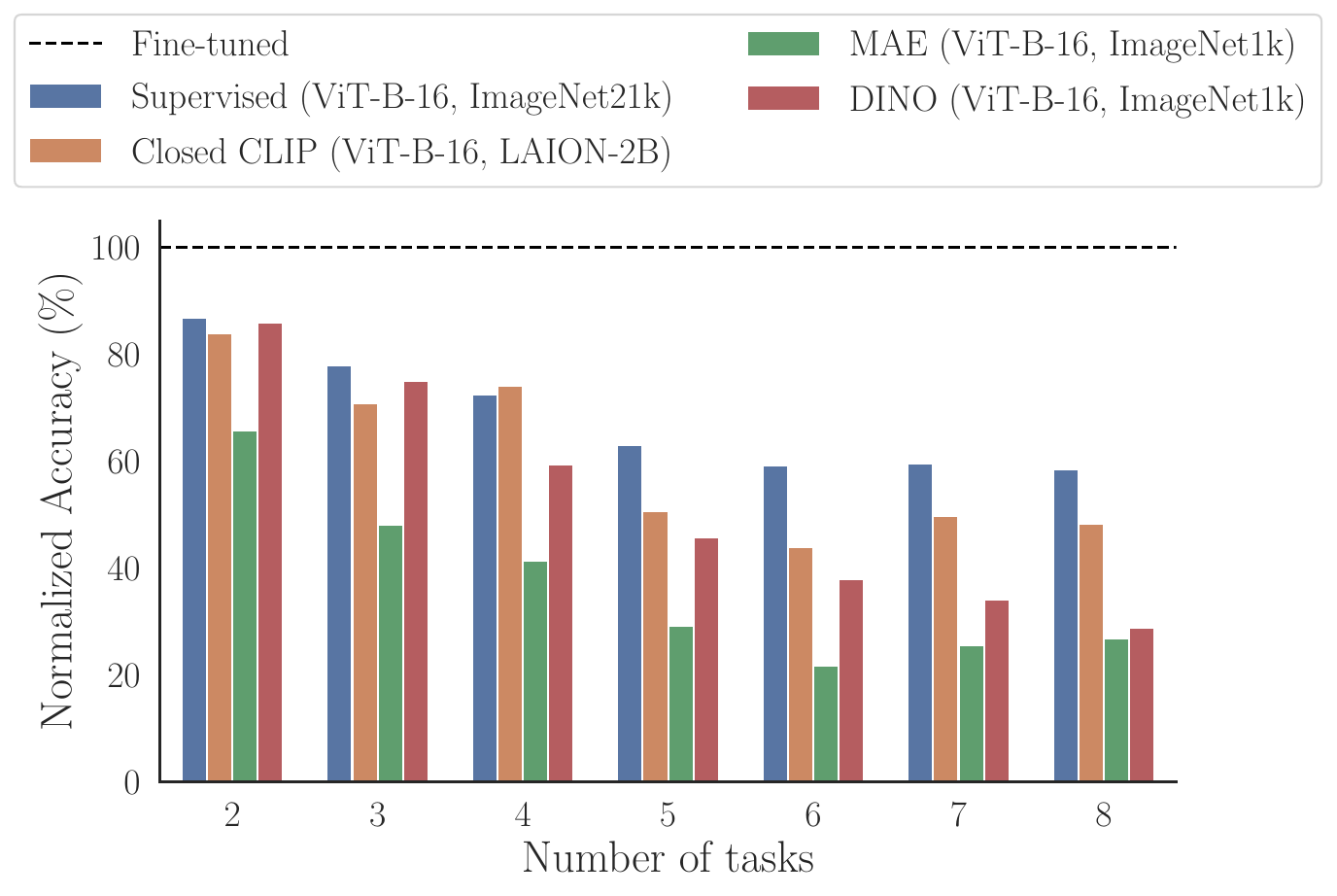}}
\vskip -0.1in
\caption{Task addition performance for all pre-training settings under a varying number of tasks.}
\label{vary_all}
\end{center}
\end{figure}

\begin{figure}[ht]
\vskip 0.2in
\begin{center}
\centerline{\includegraphics[width=300pt]{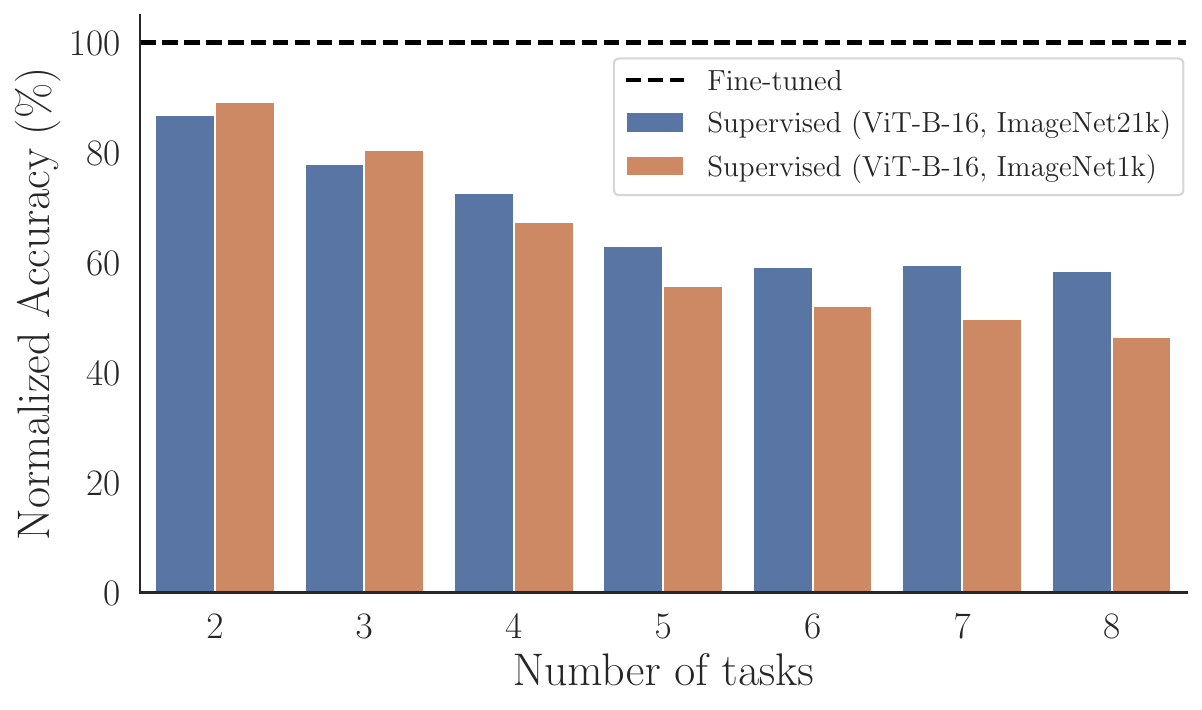}}
\caption{Normalized task addition accuracy for Supervised ViT-B-16 pre-trained on ImageNet1k vs. ImageNet21k under a varying number of tasks. }
\label{vary_data_size}
\end{center}
\vskip -0.2in
\end{figure}

\begin{figure}[ht]
\vskip 0.2in
\begin{center}
\centerline{\includegraphics[width=300pt]{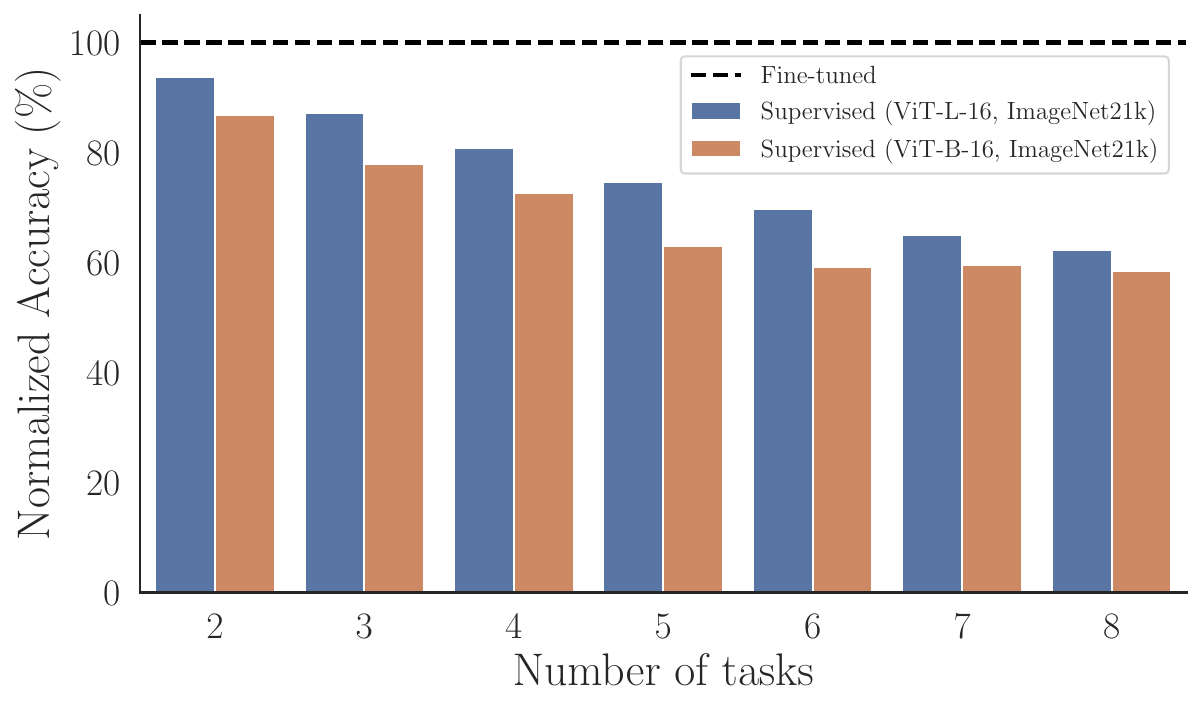}}
\caption{Normalized task addition accuracy for a Supervised ViT-B-16 vs. ViT-L-16 pre-trained on ImageNet21k under a varying number of tasks. }
\label{vary_model_scale}
\end{center}
\vskip -0.2in
\end{figure}

In general, in Figure \ref{vary_all}, we observe that most of our models, except for MAE, maintain high normalized task addition accuracy for 2-3 tasks, and this performance drops as we add more tasks. We also show, in Figures \ref{vary_data_size} and \ref{vary_model_scale} respectively, that increasing the model scale from 86M parameters (ViT-B-16) to 300M (ViT-L-16) and the pre-training data size from 1M samples (ImageNet1k) to 14M (ImageNet21k) both yield small asymptotic improvements in task addition accuracy.

\end{document}